\documentclass{article}

\usepackage[latin9]{inputenc}
\usepackage{units}
\usepackage{amsmath}
\usepackage{amssymb}

\usepackage{graphicx} 
\usepackage{subfigure}

\usepackage{natbib,natbibspacing}

\usepackage{algorithmic}

\usepackage{hyperref}

\usepackage{multirow}

\usepackage{algorithm2e}
\usepackage{bm}
\usepackage{overpic}
\usepackage{multirow}

\usepackage{textcomp}

\usepackage{graphicx} 
\usepackage{subfigure}

\newcommand{\mat}[1]{\ensuremath{\mathbf{#1}}}

\def\datam{\mat{X}}
\def\datav{\mat{x}}
\def\data{x}

\def\coefv{\mat{z}}

\def\reals{\ensuremath{\mathbb{R}}}
\def\naturals{\ensuremath{\mathbb{N}}}

\newcommand{\norm}[1]{\ensuremath{\left\|#1\right\|}}    

\newcommand{\argmin}[1]{\underset{#1}{\operatorname{argmin}}}

\newcommand{\mypara}[1]{{\noindent \bf{#1:} }}

\newcommand{\bb}[1]{\bm{\mathrm{#1}}}

\def\Tr{\mathrm{T}}

\title{Learning Robust Low-Rank Representations}

\author{
Pablo Sprechmann\thanks{Authors' email addresses: \texttt{pablo.sprechmann@duke.edu}, \texttt{bron@eng.tau.ac.il}, and \texttt{guillermo.sapiro@duke.edu}. }\\
Dept. of Elec. and Comp. Eng.\\
Duke University\\
\and
Alexander M. Bronstein \\
School of Elec. Eng.\\
Tel Aviv University\\
\and
Guillermo Sapiro\\
Dept. of Elec. and Comp. Eng.\\`
Duke University\\
}

\begin{document}

\maketitle

\begin{abstract}

In this paper we present a comprehensive framework for learning robust
low-rank representations by combining and extending recent ideas for
learning fast sparse coding regressors with structured non-convex
optimization techniques. This approach connects robust principal
component analysis (RPCA) with dictionary learning techniques and
allows its approximation via trainable encoders. We propose an
efficient feed-forward architecture derived from an optimization
algorithm designed to exactly solve robust low dimensional
projections. This architecture, in combination with different training
objective functions, allows the regressors to be used as online
approximants of the exact offline RPCA problem or as RPCA-based neural
networks. Simple modifications of these encoders can handle
challenging extensions, such as the inclusion of geometric data transformations.
We present several examples with real data from image, audio, and video
processing. When used to approximate RPCA, our basic implementation
shows several orders of magnitude speedup compared to the exact
solvers with almost no performance degradation. We show the strength
of the inclusion of learning to the RPCA approach on a music source separation
application, where the encoders outperform the exact RPCA algorithms,
which are already reported to produce state-of-the-art results on a benchmark
database. Our preliminary implementation on an iPad shows
faster-than-real-time performance
with minimal latency.

\end{abstract}

\section{Introduction}
\label{sec.introduction}

Principal component analysis (PCA) is the most widely used statistical technique for dimensionality
reduction, with applications ranging from machine learning and computer vision to signal processing and bioinformatics, just to mention a few.
Given a data matrix $\datam \in \reals^{m \times n}$ (each column of $\datam$ is an $m$-dimensional data vector),
it is decomposed as $\datam = \bb{L} + \bb{N}$,
where $\bb{L}$ is a low rank matrix and $\bb{N}$ is a perturbation matrix.
%
PCA is known to produce very good results when the perturbation is small \cite{pca}. However, its performance is highly sensitive to the presence of samples not following the model; even a single outlier in the data matrix $\datam$ can render the estimation of the low rank component arbitrarily far from the true matrix $\bb{L}$.
This motivated an important amount of work dedicated to robustifying PCA, see \cite{Torre03aframework,Candes2011-JACM} for recent work and references therein for previous results.
In a series of recent works \cite{Candes2011-JACM,NIPS2010_0849}, a very elegant solution to this problem was developed, in which the low rank matrix is determined as the minimizer of a convex program. The basic idea is to add a new term in the decomposition to account for the presence of outliers,
$\datam = \bb{L} + \bb{N} + \bb{O}$, where $\bb{O}$ is an error matrix with a sparse number of non-zero coefficients with arbitrarily large magnitude.
\emph{Robust} PCA is then obtained by solving
\begin{equation}
\min_{\bb{L}, \bb{O}\in\reals^{m\times n}}
\norm{\bb{L}}_\ast + \lambda \norm{\bb{O}}_1 \quad \quad \textrm{s.t.} \quad \norm{\datam - \bb{L} - \bb{O}}^2_F \leq \epsilon,
\label{ec.convex}
\end{equation}
where $\norm{\bb{L}}_\ast$ denotes the matrix nuclear norm, defined as the sum of the singular values of $\bb{L}$ (the convex surrogate of the rank), $\lambda$ is a positive scalar parameter controlling the sparsity level in the outliers, and $\epsilon$ is a parameter controlling the error of the approximation.
In the noiseless setting, the constraint is often substituted by the equality $\datam = \bb{L} + \bb{O}$ \cite{Candes2011-JACM}.

This particular formulation of robust PCA has attracted significant
interest in the machine learning, computer vision, and signal processing communities, and
was successfully used in applications such as face recognition and
modeling \cite{Wagner2011-PAMI,Peng2011-PAMI}, background modeling \cite{Zhou2010,realtimeRPCA}, large scale image tag
transduction \cite{MuDYY11}, and audio source separation \cite{huang12}. A challenge often
encountered in modern applications is that the flow of new input data
is permanent. Then, the robust low rank model needs to be adapted
constantly since the principal directions can change over time,
calling for developing efficient online techniques \cite{realtimeRPCA,TanCM11,mateos-2011,grouse}.

Significant amount of effort has been devoted to developing
optimization algorithms for efficiently solving (2) and its noiseless
formulation. First-order techniques based on proximal methods \cite{Candes2011-JACM,CaiCS10,Lin2009-pp}, and augmented Lagrangian approaches \cite{MaGC11} have shown to be fast and effective when the data size is moderate. More recent efforts proposed
methods using random projections with drastically reduced scale \cite{MuDYY11},
or by decomposing (2) into a non-convex structured problem of
significantly reduced size \cite{Recht:2011wv,mateos-2011}. Despite the permanent progress
reported in the literature, state-of-the-art algorithms for solving
this problem still have prohibitive complexity and latency for real-time processing.

In the sparse coding domain, a very similar situation was encountered
a few years ago. Techniques based on sparse representations in
learned over-complete dictionaries produced outstanding results in
various computer vision and signal processing problems, but they are often
prohibitively costly for real-time computation. This motivated significant effort in the deep learning
community aiming at overcoming this problem. On one hand, several works
concentrated on proposing systems capable of producing sparse codes,
aiming to bring the success of the exact sparse coding algorithms to
extremely efficient deep learning schemes, e.g., \cite{RanzatoHBL07,autoencoders}. In a different approach,
several works proposed learning non-linear
regressors capable of producing good approximations of the true sparse
codes in a fixed amount of time \cite{jarrett2009best,kavukcuoglu2010fast}.
The insightful work in \cite{LecunNN}
introduced an approach in which the regressors are multilayer
artificial neural networks with an architecture inspired by first
order optimization algorithms for solving sparse coding problems.
These regressors are trained to minimize the mean squared error
between the predicted and exact codes over a given training set, and
produced high quality approximations of sparse codes for vectors
following the same distribution as the training sample.

Motivated by the latter approach, in this paper we propose to extend these ideas to
the RPCA context. We propose to design regressors
capable of approximating online RPCA in a very fast way. To
the best of our knowledge, this type of encoders have never been developed
before. We follow \cite{LecunNN} and base the
architecture of the encoders on the iterations of exact RPCA algorithms.
However, unlike the standard sparse coding setting, the
exact first order RPCA algorithms cannot be used directly, as each iteration involves
SVD. As a remedy, we use an algorithm inspired by the non-convex optimization techniques proposed
in \cite{Recht:2011wv}. Our RPCA encoders are learned by minimizing various carefully
chosen objective functions that allow their use in several different
contexts as explained in the sequel.

Learning encoders to approximate RPCA becomes particularly relevant
when the low rank model has to be re-computed or updated constantly
throughout time. We propose a training objective function that allows
the encoders to be trained in an online manner on the very same data
vectors fed to them. This also makes the fast encoders no more restricted to work with a specific distribution of input vectors known
\emph{a priori} (limitation existing, for example, in \cite{LecunNN}), and removes the need to run the exact algorithms
beforehand. This approach is related to the sparse autoencoders
\cite{autoencoders}, as will be further discussed in Section~\ref{sec.online}.

Several applications of  RPCA rely on the critical assumption that the
given input vectors  are \emph{aligned} with respect to a group of
geometric transformations \cite{Peng2011-PAMI}. The state-of-the art techniques for
addressing this problem involve the computation of several RPCA
problems by changing the individual transformations applied to each
input data vector. The differentiability of the proposed encoders with
respect to the input, output and training data allows a very simple
incorporation of geometric transformations.  We propose a learning
setting for that important case as well.

Finally, in many applications RPCA is applied to signals not
exactly following the low rank model with sparse additive outliers. A
clear example is the problem of separating the leading singing voice
from the musical background from a monaural recording, as detailed in Section~\ref{sec.experiments}.
In \cite{huang12}, the authors
obtained state-of-the-art results in this problem using RPCA in the
time-frequency domain by modeling the repetitive structure of the
accompaniment as a low-rank linear model and the singing voice as
sparse outliers. It is clear, however, that using a richer model for
representing the singing voice (e.g., the harmonic structure makes the
voice patterns highly structured) would produce a better performance.
We propose to fill in the gap between the RPCA model and the real
signals by incorporating learning, that is, by changing the training
objective function of our RPCA encoders so that they approximate the
desired source separation.
Experimental evaluation shows the benefit
of this approach, which serves as an illustration of the use of our proposed framework for different objective functions and tasks other than reconstruction.

The rest of the paper is organized as follows:
In Section~\ref{sec.sparse.models}, we present our approach to robust PCA
and discus exact optimization algorithms to solve it. In Section~\ref{sec.NN}, we introduce the new robust encoders and the new objective functions used for their training.
We also discuss the online setting and the possibility to incorporate geometric transformations. In Section~\ref{sec.experiments}, we present several experimental results.
Conclusions are drawn in Section~\ref{sec.concl}.


\section{Online RPCA via non-convex factorization}
\label{sec.sparse.models}

\label{sec.nonconvex.rpca}

In this paper we tackle the RPCA problem by solving the unconstrained optimization problem
\begin{equation}
\min_{\bb{L}, \bb{O}\in\reals^{m\times n}}
\frac{1}{2} \norm{\datam - \bb{L} - \bb{O}}^2_F + \lambda_\ast \norm{\bb{L}}_\ast + \lambda \norm{\bb{O}}_1.
\label{ec.convex.unconstrained}
\end{equation}
This formulation is equivalent to \eqref{ec.convex} in the sense that for every $\epsilon>0$ one can
find a $\lambda_\ast>0$ such that \eqref{ec.convex} and \eqref{ec.convex.unconstrained} admit the same solutions.

As the $\ell_1$-norm encourages sparsity with vectors, the nuclear norm promotes low rank in matrices.
\cite{RechtFazelParrilo2007} showed that the nuclear norm of a matrix of $\mathrm{rank}(\bb{L}) \le q$ can be reformulated as a penalty over all possible factorizations
\begin{equation}
\norm{\bb{L}}_\ast =  \min_{\bb{U}, \bb{S}}
\frac12 \norm{\bb{U}}_F^2 + \frac12 \norm{\bb{S}}_F^2 \quad \quad \textrm{s.t.} \quad \bb{U}\bb{S} = \bb{L},
\label{ec.nuclear.norm.eq}
\end{equation}
$\bb{U} \in \reals^{m \times q}$, $\bb{S} \in \reals^{q \times n}$.
The minimum is achieved by the SVD: if $\bb{L} = \bb{U}_L\bb{\Sigma}\bb{V}_L^T$ then the minimum of \eqref{ec.nuclear.norm.eq} is $\bb{U}=\bb{U}_L\bb{\Sigma}^{\frac12}$ and $\bb{S}=\bb{\Sigma}^{\frac12}\bb{V}_L$. This factorization has been recently exploited in parallel processing across multiple processors to produce state-of-the-art algorithms for matrix completion problems \cite{Recht:2011wv}, as well as an alternative approach to robustifying PCA in \cite{mateos-2011}.

In \eqref{ec.convex.unconstrained}, neither the rank of $\bb{L}$ nor the level of sparsity in $\bb{O}$ are assumed known \emph{a priori}.
However, in many applications, it is a reasonable to have a rough upper bound of the rank, say $\textrm{rank}(\bb{L})\leq q$.
Combining this with \eqref{ec.nuclear.norm.eq}, we can reformulate \eqref{ec.convex.unconstrained} as
\begin{equation}
\min_{\bb{U}, \bb{S}, \bb{O}}
\frac{1}{2} \norm{\datam - \bb{U}\bb{S} - \bb{O}}^2_F + \frac{\lambda_\ast}{2} ( \norm{\bb{U}}_F^2 + \norm{\bb{S}}_F^2 ) + \lambda \norm{\bb{O}}_1,
\label{ec.convex.unconstrained.US}
\end{equation}
with $\bb{U}\in\reals^{m\times q}$, $\bb{S}\in\reals^{q\times n}$, and $\bb{O}\in\reals^{m\times n}$. This decomposition reveals a lot of structure hidden in the problem. The low rank component can now be thought as an under-complete dictionary $\bb{U}$, with $q$ atoms, multiplied by a matrix $\bb{S}$ containing in its columns the corresponding coefficients for each data vector in $\datam$. This interpretation brings our problem close to that of dictionary learning in the sparse modeling domain \cite{mairal2009online}.

While this new factorized formulation drastically reduces the number of optimization variables from $2nm$ to $q(n+m)$, problem \eqref{ec.convex.unconstrained.US} is no longer convex. Fortunately, it can be shown that
any stationary point of \eqref{ec.convex.unconstrained.US}, $\{\bb{U},\bb{S},\bb{O}\}$, satisfying $||\datam - \bb{U}\bb{S} - \bb{O}||^2_2\leq \lambda_\ast$ is an optimal solution of \eqref{ec.convex.unconstrained.US} \cite{morteza}.
Thus, problem \eqref{ec.convex.unconstrained.US} can be solved using an alternating minimization or block coordinate scheme, in which the cost function is
minimized with respect to each individual optimization variable while keeping the other ones fixed, without the risk of falling into a stationary point that may
not be globally optimal.
This will be exploited to design our fast encoders.

\subsection{Robust low dimensional projections}
\label{sec.robust.projection}

Let us assume that we have already learned a low dimensional model, $\bb{U}\in \reals^{m\times q}$, from some data $\datam\approx \bb{U}\bb{S}+ \bb{O}\in \reals^{m\times n}$. Suppose that we are given a new input vector $\bb{\hat{x}}\in \reals^m$ drawn from the same distribution as $\datam$.
Then $\bb{x}$ can be decomposed as $\bb{x} =\bb{U}\bb{s}+\bb{n}+\bb{o}$, where $\bb{U}\bb{s}$ represents the low dimensional component, $\bb{n}$ is a small perturbation and $\bb{o}$ is a sparse outlier vector. We propose to do it by extending \eqref{ec.convex.unconstrained.US}
\begin{equation}
\min_{\bb{s}\in \reals^{q}, \bb{o}\in \reals^m} \frac{1}{2} \norm{\bb{x} - \bb{U} \bb{s} - \bb{o}}^2_2 + \frac{\lambda_\ast}{2} \norm{\bb{s}}_2^2  + \lambda \norm{\bb{o}}_1.
\label{ec.robust.projection}
\end{equation}
Unlike dictionary learning problems \cite{mairal2009online}, here the columns of the dictionary $\bb{U}$ are not constrained to have unit norm. In fact, the differences in the norms of the different atoms play a crucial role in the estimation weighting the relevance of the atoms in the low dimensional distribution and appearing in the objective function as the quadratic term $\norm{\bb{s}}_2^2$. To give some further intuition we analyze the program \eqref{ec.robust.projection} and its relation with the possible solutions of \eqref{ec.convex.unconstrained.US}.
As discussed in the previous section, if the upper bound $q$ for the rank of the true low dimensional model is correct, any pair of matrices $\{\bb{U},\bb{S} \}$ found as a stationary point of \eqref{ec.convex.unconstrained.US} and satisfying $||\datam - \bb{U}\bb{S} - \bb{O}||^2_2\leq \lambda_\ast$, is guaranteed to satisfy $\bb{L}= \bb{U}\bb{S}$.
For simplicity in the notation and without loss of generality, in the sequel we assume that the rank of $\bb{L}$ is exactly $q$.
Then, the solution of program \eqref{ec.robust.projection}, $\{\bb{s},\bb{o}\}$, satisfies $\bb{U}\bb{s}= \bb{\tilde{U}}\bb{\tilde{s}}_2$ and $\bb{o}=\bb{\tilde{o}}$,
where $\{\bb{\tilde{s}},\bb{\tilde{o}}\}$ is the solution obtained by substituting $\bb{U}$ by $\bb{\tilde{U}}=\bb{U}_L\bb{\Sigma}^{\frac12}$ in \eqref{ec.robust.projection}.
Applying the change of coordinates $\bb{s} = \bb{\Sigma}^{-\frac12}\bb{w}$, this new problem can be written as
\begin{equation}
\min_{\bb{w}\in \reals^{q}, \bb{o}\in \reals^m} \frac{1}{2} \norm{\bb{\hat{x}} - \bb{U}_L\bb{w} - \bb{o}}^2_2 + \frac{\lambda_\ast}{2} \sum\limits_{i=1}^{q} \frac{w_i^2}{\sigma_i}+ \lambda \norm{\bb{o}}_1,
\label{ec.robust.projection.weights}
\end{equation}
where the $w_i$'s are the individual coefficients of $\bb{w}$ and the $\sigma_i$'s are the singular values of $\bb{L}$, $\bb{\Sigma} = \textrm{diag}(\sigma_1,\ldots,\sigma_{q})$.
Note that $\bb{U} \bb{s} = \bb{U}_L \bb{w}$. The second term in the cost in \eqref{ec.robust.projection.weights} acts as a regularizer that encourages the use of the coefficient of $\bb{w}$ corresponding to the dominant directions (larger singular values) of $\bb{L}$.

The robust low dimensional projection \eqref{ec.robust.projection} is a convex program that can be solved using several methods. We are interested in choosing an optimization algorithm that can be further used to define the architecture of trainable encoders for simultaneously estimating $\bb{s}$ and $\bb{o}$. With this in mind, we choose to use the alternating minimization scheme, described in Algorithm~\ref{alg:so}. The solution of \eqref{ec.convex.unconstrained.US.online} is given by
$\bb{s} = (\bb{U}^\Tr\bb{U}-\lambda_\ast \bb{I} )^{-1}\bb{U}^\Tr (\bb{x}_t - \bb{o})$
and
$\bb{o} = \bb{\pi}_{\lambda}(\bb{\hat{x}} - \bb{U}\bb{s})$,
when fixing $\bb{o}$ and $\bb{s}$ respectively. Here $\bb{\pi}_{\bb{\lambda}}$ is the scalar soft-thresholding operator with parameter $\bb{\lambda}\in \reals^m$,
which applies a soft-threshold $\lambda_i$ to each component of the input vector. In this case, $\bb{\lambda}=\lambda\bb{1}$.

\begin{algorithm}[tb]

\SetKwInOut{Input}{input}\SetKwInOut{Output}{output}

\Input{Data $\hat{\datav}$, dictionary $\bb{U}$, parameters $\lambda_\ast$ and $\lambda$.}

\Output{Coefficient vector $\bb{s}$ and outlier vector $\bb{o}$.}

Define $\bb{H} =  (\bb{U}^T\bb{U}-\lambda_\ast \bb{I} )^{-1}$ and $\bb{W} =\bb{U}\bb{H}\bb{U}^T$, $\bb{\lambda}=\lambda\bb{1}$.

Initialize $\bb{y} = \bb{0}$ and $\bb{b} = (\bb{I} - \bb{W}) \datav$.

\Repeat{until convergence}
{

$\bb{o} = \bb{\pi}_{\bb{\lambda}}(\bb{b})$

$\bb{b} = \bb{b} + \bb{W} (\bb{o} - \bb{y})$

$\bb{y} = \bb{o}$
}
Output $\bb{o}$ and $\bb{s} = \bb{H}(\datav - \bb{o})$.
\vspace{5mm}
\caption{Alternating minimization scheme for solving \eqref{ec.robust.projection}.\label{alg:so}
\vspace{-2ex}
}

\end{algorithm}

\subsection{Online RPCA}
\label{sec.online.rpca}

In Section~\ref{sec.nonconvex.rpca} we assumed that the entire data matrix $\datam$ was available
\emph{a priori}. We now address the case when the data samples $\{\bb{x}_t\}_{t\in \naturals}$, $\bb{x}_t\in \reals^{m}$,  arrive sequentially;
the index $t$ should be understood as a temporal variable. Online RPCA aims at estimating and refining the model as the data comes in.
The need for online algorithm appears naturally in a various applications, e.g., when large volumes of data are
permanently generated over time. Other applications aim at estimating models
for dynamic data constantly changing over time. Finally, online learning
has also been extensively used when the available
training data are simply too large to be handled together \cite{mairal2009online}.

We propose to address online RPCA extending the approach
presented in Section~\ref{sec.sparse.models}.
An  alternating minimization algorithm for solving the online counterpart of \eqref{ec.convex.unconstrained.US} goes as follows:
When a new data vector $\bb{x}_t$ is received, we first obtain its representation $\{\bb{s}_t,\bb{o}_t\}$ given the current model estimate, $\bb{U}_{t-1}$.
\begin{eqnarray}
\{\bb{s}_t, \bb{o}_t\} &=&\argmin{\bb{s}, \bb{o}}
\frac{1}{2} \norm{\bb{x}_t - \bb{U}_{t-1}\bb{s} - \bb{o}}^2_2 + \frac{\lambda_\ast}{2} \norm{\bb{s}}_2^2  + \lambda \norm{\bb{o}}_1.
\label{ec.convex.unconstrained.US.online}
\end{eqnarray}
Then, we update the model using the projections, $\{\bb{s}_j\}_{j\leq t}$ and $\{\bb{o}_j\}_{j\leq t}$, computed during the previous steps of the algorithm,
\begin{eqnarray}
\bb{U}_t &=& \argmin{\bb{U}} \sum\limits_{j=1}^{t} \beta_j \left(\frac{1}{2} \norm{\bb{x}_j - \bb{U}\bb{s}_j - \bb{o}_j}^2_2 + \frac{\lambda_\ast}{2} \norm{\bb{U}}_F^2 \right),
\label{ec.convex.unconstrained.US.online.dict}
\end{eqnarray}
where  $\beta_j\in [0,1]$ is a forgetting factor that can be added to rescale older information so that newer estimates $\{\bb{s}_j, \bb{o}_j\}$ have more weight.

Problem \eqref{ec.convex.unconstrained.US.online} is identical to \eqref{ec.robust.projection} and can be solved using Algorithm~\ref{alg:so} setting $\bb{U}=\bb{U}_{t-1}$ and $\hat{\datav} = \data_t$. There are two major approaches to solving \eqref{ec.convex.unconstrained.US.online.dict}.
The first one is to solve it recursively from previous estimates, using strategies such as block-coordinate descent methods with warm restarts \cite{mairal2009online} or recursive least squares \cite{mateos-2011}. The other option is to directly solve the system of equations
\begin{equation}
\bb{U}_{t}\left(\sum\limits_{j=1}^{t} \beta_j \bb{s}_j \bb{s}_j^\Tr + \lambda_\ast \bb{I} \right) = \sum\limits_{j=1}^{t} \beta_j (\bb{x}_j-\bb{o}_j) \bb{s}_j^\Tr.
\label{ec.linear.U}
\end{equation}
While the recursive strategies have, in general, lower computational complexity, in particular for large scale problems, they require more storage.
The choice of the dictionary update strategy is, therefore, application-dependent.

\section{Online RPCA via fast trainable encoders}
\label{sec.NN}

As mentioned in the Introduction, one of the main contributions of the present paper is the construction of trainable regressors
capable of approximating the solution of \eqref{ec.robust.projection} for a given fixed dictionary $\bb{U}$ (the latter will be updated as well as shown in the sequel).
The main idea is to build a parametric regressor $\bb{z} = (\bb{s},\bb{o}) = \bb{h}(\bb{x},\bb{\Theta})$, with some set of parameters, collectively denoted as $\bb{\Theta}$.
Thus, we need to define an architecture for $\bb{h}$ and a learning algorithm in order to determine $\bb{\Theta}$.

Following the fast sparse coding methods in \cite{LecunNN,ICML}, we propose to use feed-forward multi-layer architecture where each layer implements a single iteration of an exact solver of the problem.
In this case we use the tailored alternating minimization scheme described in Algorithm~\ref{alg:so}. The parameters of the network are the matrices $\bb{W}$ and $\bb{H}$ and the thresholds $\bb{\lambda}$ (extra flexibility is obtained by learning different thresholds $\lambda_i$ for each component). The encoder architecture is depicted in Figure~2 in the supplementary material.
Each layer essentially consists of the nonlinear thresholding operator $\bb{\pi}_{\bb{\lambda}}$ followed by a linear operation $\bb{W}$. The network parameters are initialized as in Algorithm~\ref{alg:so}.

As a learning strategy, we propose to select the set of parameters $\bb{\Theta}$ that minimizes the loss function,
\begin{eqnarray}
\mathcal{L}(\bb{\Theta}) &=& \frac{1}{n}\sum_{j=1}^n L(\bb{\Theta}, \bb{x}_j)
\label{eq:NNtraining}
\end{eqnarray}
on a training set $\bb{X}=\{\bb{x}_1,\dots,\bb{x}_n\}$. Here, $L(\bb{\Theta}, \bb{x}_j)$ is a function that measures the goodness of the code $\bb{z}_j =\bb{h}(\bb{x}_j,\bb{\Theta})$ produced for the data point $\bb{x}_j$. We will discuss bellow several different options for choosing $L$. The selection of the objective function $L$ sets the type of regressor that we are going to obtain and this is clearly application dependent. 

One of the most straightforward choices is to use $L(\bb{\Theta}, \bb{x}_j) = \| \bb{z}_j - \bb{z}^\ast_j \|$, with $\bb{z}^\ast_j = (\bb{s}^\ast_j,\bb{o}^\ast_j)$ being the $j$-th columns of the decomposition of the data $\bb{X} = (\bb{x}_1,\dots,\bb{x}_n)$ into $\bb{X} = \bb{U}\bb{S}^\ast + \bb{O}^\ast$ by the exact RPCA. This essentially trains the encoder to approximate the exact solution of the RPCA problem.
In other applications, the data may not completely adhere to the assumptions of the RPCA model, and the exact solution is, therefore, not necessarily the best one.
This is the case in the source separation problem discussed in the introduction, where RPCA gives a very good separation of the spectrally sparse singing voice and
repetitive low-rank background accompaniment, yet the obtained signals are still not equal to the true voice and background tracks.
In this case, one could use a collection of clean voice and background tracks, $\{\bb{s}^\ast_j$\} and $\{\bb{o}^\ast_j\}$ respectively,
to supervise the training, often achieving better separation results.
Other choices of the loss function are discussed in the sequel.

%

We perform the minimization of a loss function $\mathcal{L}(\bb{\Theta})$ with respect to $\bb{\Theta}$ using a stochastic gradient descent, as in \cite{LecunNN}.
Specifically, we iteratively select a random subset of $\mathcal{X}$ and then update the network parameters
as $\bb{\Theta} \leftarrow \bb{\Theta} - \mu \frac{\partial \mathcal{L}(\bb{\Theta}) }{\partial \bb{\Theta}}$,
where $\mu$ is a decaying step, repeating the process until convergence. This requires the computation of the (sub)gradients $d\mathcal{L}(\bb{\Theta}, \bb{x}_t ) / d\bb{\Theta}$, which is achieved by a back-propagation procedure.
%

\subsection{Online learning}
\label{sec.online}

The robust projection \eqref{ec.robust.projection} can be viewed as a mapping between a data vector $\bb{x}$ and the corresponding pair $\bb{z} = \{\bb{s},\bb{o}\}$ minimizing the cost function,
\begin{eqnarray}
f(\bb{x},\bb{z}) = \frac{1}{2} \norm{\bb{x} - \bb{U} \bb{s} - \bb{o}}^2_2 + \frac{\lambda_\ast}{2} \norm{\bb{s}}_2^2  + \lambda \norm{\bb{o}}_1.
\label{eq:sparsity-objective}
\end{eqnarray}
This objective is trusted as an indication of the decomposition quality as explained in Section~\ref{sec.robust.projection}. Then, the network can be trained to minimize the ensemble average of $f$ on a training set with $\coefv = \mathrm{arg}\min f(\datav, \coefv)$ replaced by $\coefv = \bb{h}(\datav, \bb{\Theta})$. This results in the training objective \eqref{eq:NNtraining} by selecting $L_{\textrm{On}}(\bb{\Theta}, \bb{x}_j) = f(\datav_j, \bb{h}(\datav_j, \bb{\Theta}))$ and
adding a forgetting factor $\beta_j$ as described in Section~\ref{sec.online.rpca}.

When the training of the regressors can be done online, one can further consider the online adaptation of the dictionary.
This can be done simply by treating $\bb{U}$ as another
optimization variable in the training, and minimizing $\mathcal{L}$ with respect to both $\bb{U}$ and the network parameters, alternating between network training and dictionary
update iterations. In this setting, the model adaptation is equivalent to \eqref{ec.convex.unconstrained.US.online.dict}. This essentially extends the proposed framework into a full-featured online RPCA encoder, trained on the very same data fed to it for robust low dimensional projections.
In this setting, our regressors can be interpreted as an online trainable sparse auto-encoder \cite{autoencoders} with a multi-layer non-linear encoder and simple linear decoder. The higher complexity of the proposed architecture in the encoder allows the system to produce accurate estimates of true structured sparse codes.

\subsection{Geometric transformations}

The underlying model used in RPCA relies on the critical assumption that the given input vectors $\datam$
are ``aligned'' with respect to each other. While this assumption holds for many applications (i.e., background subtraction with static cameras), it does not apply in all cases. The canonical example is face modeling, where the low dimensional model only holds if the facial images are pixel-wise aligned \cite{Peng2011-PAMI}.
Even small misalignments can break the structure in the data, the representation quickly degrades as the rank of the low dimensional component $\bb{U}\bb{S}$ increases and the matrix of outliers $\bb{O}$ loses its sparsity.

This challenging problem has been recently studied in the literature. In \cite{iccvK-S11} the authors propose a pre-processing strategy to align the training images. In \cite{Peng2011-PAMI}, the authors simultaneously align the input images and solve RPCA with a sequence of convex problems.

Following \cite{Peng2011-PAMI}, we propose to incorporate the optimization over geometric transformations of the input data into the representation problem. Then, the optimization problem \eqref{ec.convex.unconstrained.US} becomes
\begin{equation}
\min_{\bb{U}, \bb{S}, \bb{O},\bb{\alpha}}
\frac{1}{2} \norm{\bb{T}_{\bb{\alpha}}(\bb{X})- \bb{U}\bb{S} - \bb{O}}^2_F + \frac{\lambda_\ast}{2} ( \norm{\bb{U}}_F^2 + \norm{\bb{S}}_F^2 ) + \lambda \norm{\bb{O}}_1,
\label{ec.convex.unconstrained.US.trans}
\end{equation}
where $\bb{T}_{\bb{\alpha}}$ is a parametrized operator (with a set of parameters collectively denoted as $\bb{\alpha}=[\bb{\alpha^1},\ldots,\bb{\alpha^n}]$) that applies different geometric transformation, $\bb{T}_{\bb{\alpha^i}}$, to each training vector $\bb{x}_i$. This formulation is highly non-convex and difficult to optimize.
Interestingly, the framework of trainable regressors introduced in Section~\ref{sec.online} is very well suited for producing accurate approximations of \eqref{ec.convex.unconstrained.US.trans} at very mild extra computational expenses.
We propose to use the training objective function defined in \eqref{eq:NNtraining} with
$L_{\textrm{Tr}}(\bb{\Theta}, \bb{x}_j,\bb{\alpha^j}) =f(\bb{T}_{\bb{\alpha^j}}(\datav_j), \bb{h}(\bb{T}_{\bb{\alpha^j}}(\datav_j), \bb{\Theta}))$.
The obtained regressors are conceptually very similar to the ones we had before and can still be trained in an online manner.
When a new data vector $\bb{x}_t$ arrives, we compute it's robust low rank projection by minimizing $L_{\textrm{Tr}}(\bb{\Theta}, \bb{x}_t,\bb{\alpha^t})$ over a the vector $\bb{\alpha^t}$ parameterizing the geometric transformation.
Here, $\bb{h}(\datav_t, \bb{\Theta})$ is almost everywhere differentiable with respect to their input $\bb{x}_t$, which allows to find the (sub)gradient with respect to $\bb{\alpha^t}$ by applying the chain rule. In the same way, as new data arrives,  the transformation of all the previously seen training vectors is updated through the minimization of a loss function $\mathcal{L}(\bb{\Theta},\bb{\alpha})$ with respect to $\bb{\alpha}$, following the same ideas in Section~\ref{sec.online}.
This strategy can also be used in the standard RPCA scenario, however, the representations $\bb{z} = \{\bb{o},\bb{s}\}$ themselves are minimizers of a convex problem, making the minimization with respect to $\bb{\alpha}$ cumbersome and computationally expensive.

\section{Experimental results}
\label{sec.experiments}

\begin{figure}
\vspace{-2ex}
\begin{centering}
\includegraphics[width=0.75\linewidth]{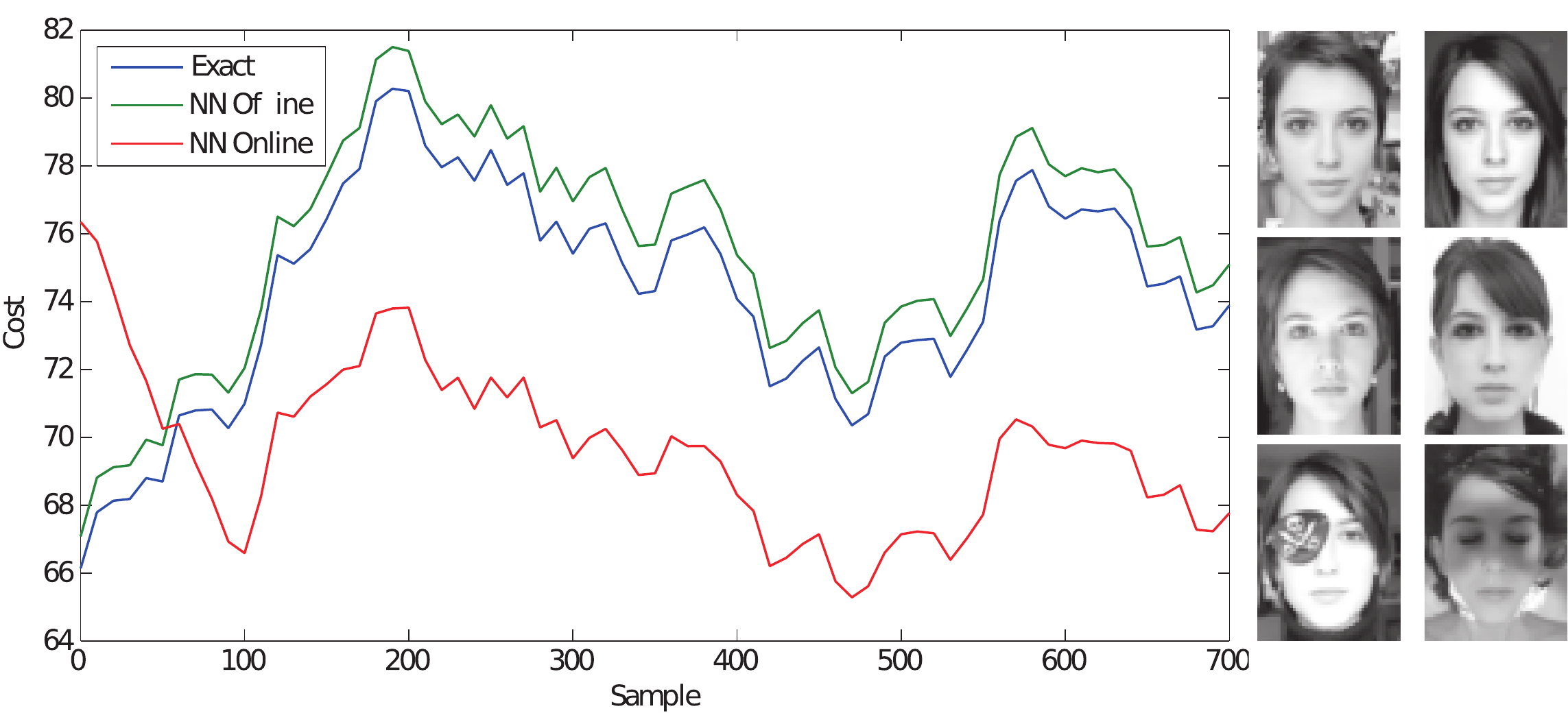}
\vspace{-1mm}
\caption{Performance, in the sense of the cost \eqref{ec.convex.unconstrained}, of the online and offline encoders on the faces sequence. Representative faces
are also shown.  \label{fig:online} }
\end{centering}
\end{figure}
\begin{figure}[t]
\includegraphics[width=\linewidth]{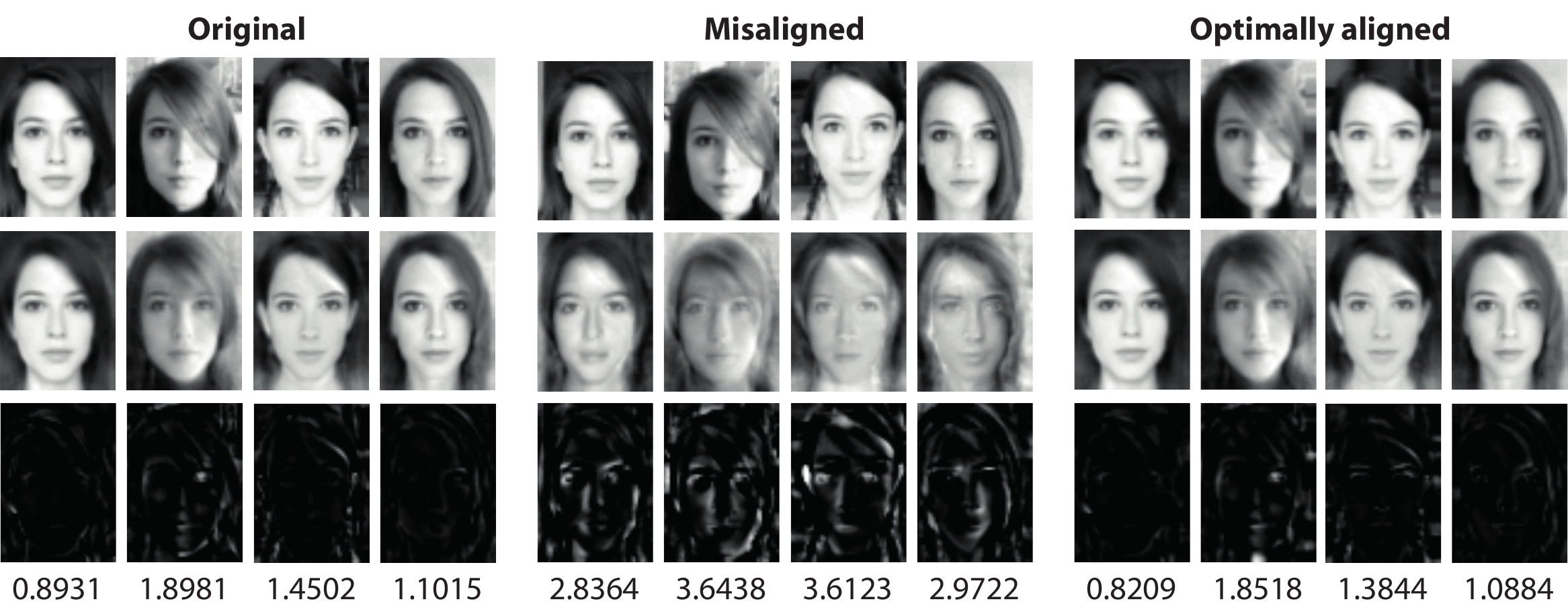}
\vspace{-1mm}
\caption{Robust PCA representation of the faces dataset in the presence of geometric transformations (misalignment).
Left group: original faces; middle group: shifted faces; right group: faces optimally realigned during encoding.
First row: reconstructed face $\bb{Us}+\bb{o}$; middle row: its low-rank approximation $\bb{Us}$; and bottom row: sparse outlier $\bb{o}$.
The $\ell_2+\ell_1$ cost is given for each representation.
 \label{fig:geometric} }
\vspace{-3ex}
\end{figure}

In what follows, we evaluate the proposed RPCA encoders on image, video, and audio data. Due to lack of space, only the essential details of the experiments are given; the reader is referred to the cited references for further details of the experimental settings that were reproduced here.
%

\begin {table}[tb]
\caption{Robust PCA representation accuracy (in the sense of the $\ell_2+\ell_1$ cost) of the faces data using different encoders. The cost for the exact encoder is $1.290$. \label{tab:faces} }
\begin{center}
\small{
\begin{tabular}{lcccc}
  \hline\hline
\multirow{2}{*}{\textbf{Encoder}}      &  \multirow{2}{*}{\textbf{Untrained}} & \multirow{2}{*}{\textbf{Supervised}} &  \multirow{2}{*}{\textbf{Unsupervised}} &  \textbf{Unsupervised} \\
                                       &                                     &                                &                                             &  \textbf{+$\bb{U}$ update} \\
\hline
\textbf{Single layer}   &   1.3355	&  1.3471    & 	1.3460 &    	1.3262 \\
\textbf{$2$ layers}   &   1.3248	&  1.3261    & 	1.3255 &    	1.3171 \\
\textbf{$10$ layers} &   1.2968	&  1.2977	&    1.2969 &    	1.2885 \\
  \hline\hline
\end{tabular}
}
\end{center}
\vspace{-3ex}
\end{table}

\mypara{Coding performance} Quality of different robust PCA encoders was evaluated on a dataset consisting of $800$ $66 \times 48$ images of a female face photographed over the timespan of $4.5$ years, roughly pose- and scale-normalized and aligned.\footnote{The original video can be found at {\footnotesize\texttt{http://www.youtube.com/watch?v=02e5EWUP5TE}}.}
Neural networks with different number of layers were trained on $500$ vectors from the faces dataset. The following training objectives were used: the $\ell_2$ discrepancy between the exact representation $\bb{s}^\ast$ and $\bb{o}^\ast$ (referred to as \emph{Supervised});
the $\ell_2+\ell_1$ objective (\ref{eq:sparsity-objective}); and the $\ell_2+\ell_1$ objective combined with the online update of the dictionary $\bb{U}$ (initial dictionary was computed using standard SVD). Parameters were set to $\lambda_\ast=0.1$ and $\lambda = 10^{-2}$.
For reference, we also report the results produced by the exact Algorithm~\ref{alg:so}, and an \emph{untrained} network with $\bb{W}$, $\bb{H}$ and $\bb{\lambda}$ initialized according to Algorithm~\ref{alg:so} (being effectively a truncated version of the algorithm).
The obtained representations are visualized in Figure~3 in the supplementary material.
Table~\ref{tab:faces} summarizes the quality of the representations in terms of the $\ell_2+\ell_1$ cost (lower numbers correspond to better quality). Note how sufficiently deep encoders with dictionary update slightly outperform the exact encoder without dictionary adaptation. Also note that using a neural network encoder to approximate the exact representations slightly degrades the $\ell_2+\ell_1$ measure compared to the untrained encoder.

\mypara{Online learning} In this experiment we evaluate the online learning capabilities of the proposed neural network encoders.
As the input data we used the time-ordered sequence of $800$ images from the faces dataset.
Online learning was performed in overlapping windows of $100$ images with a step of $10$ images. We compared the exact algorithm,
a five layer neural network encoder trained offline using the $\ell_2 + \ell_1$ objective (\emph{NN offline}), and the same encoder trained online with adaptive $\bb{U}$.
The dictionary was initialized using SVD.
Performance measured in terms of the exact cost (\ref{ec.convex.unconstrained}) is reported in Figure~\ref{fig:online}. The exact offline encoder is consistently slightly inferior to the exact algorithm. However, thanks to its capabilities to adapt to the changing data distribution, the online encoder starts outperforming the offline counterparts after a relatively brief period of initial adaptation.

\mypara{Geometric transformations} We now evaluate the representation capabilities of the proposed neural network encoder in the presence of geometric transformations. A five layer encoder was trained on $600$ images from the faces dataset. As the test set, we used the remaining $200$ faces, as well as a collection of geometrically transformed images from the same test set. Sub-pixel planar translations were used for geometric transformations.
The encoder was applied to the misaligned set, optimizing  the $\ell_2+\ell_1$ objective over the transformation parameters.
For reference, the encoder was also applied to the transformed and the untransformed test sets without performing optimization.
Examples of the obtained representations are visualized in Figure~\ref{fig:geometric}.
Note the relatively larger magnitude and the bigger active set of the sparse outlier vector $\bb{o}$ produced for the misaligned faces, and how they are re-aligned when optimiziation over the transformation is allowed. Since the original data are only approximately aligned, performing optimal alignment during encoding frequently yields lower cost compared to the plain encoding of the original data.

\begin{figure}[t]
\begin{center}
\includegraphics[width=1\linewidth]{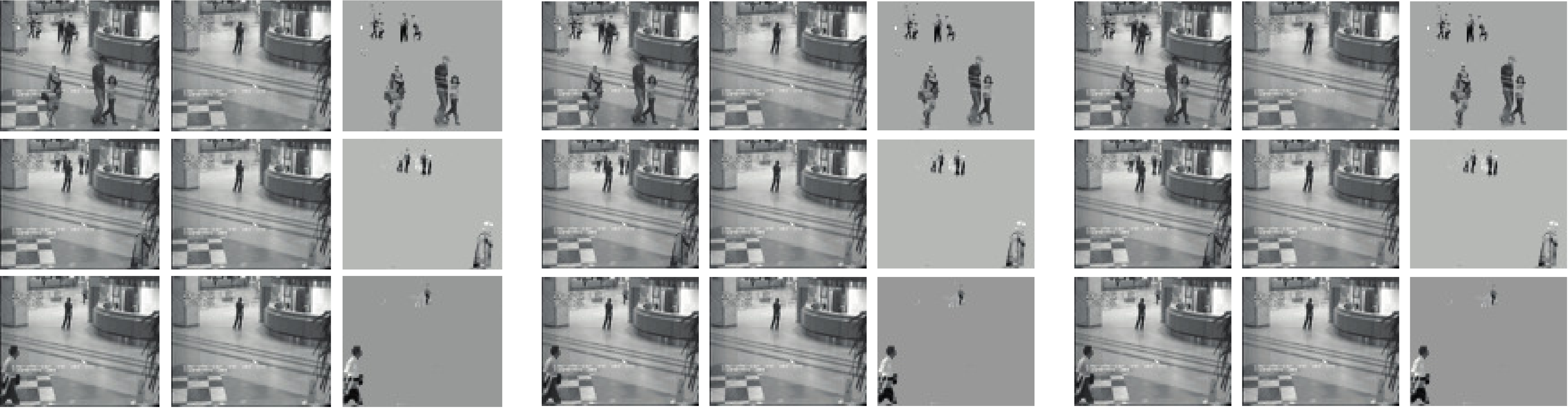}
\end{center}
\vspace{-2ex}
\caption{\small{Robust PCA representation of several frames from the surveillance sequence
obtained using the algorithm in \cite{Lin2009-pp} (left group), Algorithm~\ref{alg:so} (middle group), and a five layer neural network encoder (right group). Columns in each group are, left-to-right: the reconstructed frame $\bb{Us}+\bb{o}$, its low-rank approximation $\bb{Us}$ (background), and the sparse outlier $\bb{o}$ (foreground). Each row corresponds to a different frame.}
\label{fig:surveillance} }
\vspace{-4ex}
\end{figure}

\mypara{Video separation}  Figure~\ref{fig:surveillance} shows background and foreground separation via robust PCA on the surveillance video sequence ``\emph{Hall of a business building}'' taken from \cite{Li04statisticalmodeling}. The sequence consists of $88 \times 72$ images of an indoor scene shot by a static camera in a mall. The scene has a nearly constant background and walking people in the foreground. We used networks with five layers and $q=5$
trained to approximate the output of the exact RPCA on a subset of the frames in the sequence. Parameters were set to $\lambda_\ast=0.1$, $\lambda = 10^{-3}$.
The separation produced by the fast encoder is nearly identical to the output of the exact algorithm and to the output of the code from \cite{Lin2009-pp}, used as reference, while being considerably faster.
Our Matlab implementation with built-in GPU acceleration executed on an NVIDIA Tesla C2070 GPU propagates a frame through a single layer of the network in merely $92\mu sec$. This is several orders of magnitude faster than the commonly used iterative solver executed on the CPU.

\mypara{Audio separation} We evaluate the separation performance of the proposed methods on the MIR-1K dataset \cite{hsu2010improvement}, containing $1000$ $16$ KHz clips extraced from $110$ Chinese karaoke songs performed by $19$ amateur singers ($11$ males and $8$ females).
Each clip duration ranges from $4$ to $13$ seconds, totaling about $133$ minutes.
We reserved about $23$ minutes of audio sang by one male and one female singers (\emph{abjones} and \emph{amy}) for the purpose of training; the remaining $110$ minutes
of $17$ singers were used for testing. The voice and the music tracks were mixed linearly with equal energy.
The experimental settings closely followed that of \cite{hsu2010improvement}, to which the reader is referred for further details.
As the evaluation criteria, we used the BSS-EVAL metrics \cite{vincent2006performance}, which calculate the global \emph{normalized source-to-distortion ratio} (GNSDR),
\emph{source-to-artifacts ratio} (GSAR), \emph{source-to-interference ratio} (GSIR), and \emph{signal-to-noise ratio} (GSNR).
All networks used $20$ layers with $q=25$. The following training regimes were compared: untrained parameters initialized according to Algorithm~\ref{alg:so} (\emph{Untrained});
unsupervised learning with the objective (\ref{ec.robust.projection}) (\emph{Unsupervised});
and training supervised by the clean voice and background tracks (\emph{Supervised}).
For reference, we also give results of ideal frequency masking as well as that of two exact RPCA algorithm minimizing (\ref{ec.convex.unconstrained}) using proximal splitting,
and its noiseless version using augmented Lagrangian.
Table~\ref{tab:comparison} summarizes the obtained separation performance. While unsupervised training makes fast RPCA encoders on par
with the exact RPCA (at a fraction of the computational complexity and latency of the latter), significant improvement is achieved by using the supervised regime.
We intend to release a demo iOS application capable of performing the separation online and in real-time on a hand-held device. 

\begin {table}[tb]
\caption{\small{Performance of audio separation methods on the MIR-1K dataset.} \label{tab:comparison} }
\begin{center}
\small{
\begin{tabular}{lcccc}
  \hline\hline
\textbf{Method} &  \textbf{GNSDR}  &  \textbf{GSNR}  &  \textbf{GSAR}  & \textbf{GSIR} \\
\hline
Ideal freq. mask          & 13.48         & 5.46             & 13.65                   & 31.22  \\
ADMoM RPCA \cite{huang12} &	5.00          & 2.38             & 6.68                    & 13.76  \\
Proximal RPCA	          & 5.48          & 3.29             & 7.02                    & 13.91  \\
NN RPCA Untrained	      & 5.30          & 2.66             & 6.80                    & 13.00  \\
NN RPCA Unsupervised      & 5.62 	      & 2.87             & 6.90                    & 14.02  \\
NN RPCA Supervised        & 6.38          & 3.18             & 7.22                    & 16.47  \\
  \hline\hline
\end{tabular}
}
\end{center}
\vspace{-3ex}
\end{table} 

\section{Conclusion}
\label{sec.concl}

By combining ideas from structured non-convex optimization with multi-layer neural networks, we have developed a comprehensive framework for the online learning of robust low-rank representations in real time and capable of handling large scale applications.
The framework includes different objective functions that allow the use of the encoders to solve challenging alignment problems at almost the same computational cost.
A basic implementation already achieves several order of magnitude speedups when compared to exact solvers, opening the door for practical algorithms following the demonstrated success of robust PCA in various applications.
Finally, robust nonegative matrix factorization can be obtained using very similar architectures.

\vspace{-1ex} 

\setlength{\bibspacing}{0.2\baselineskip}

\bibliography{egbib}
\bibliographystyle{icml2012}

\end{document}